\documentclass[11pt, a4paper, logo, twocolumn]{deepmind}

\usepackage{amsmath, latexsym}
\usepackage{amsfonts}
\usepackage{mathtools}
\usepackage{ntheorem}
\usepackage{dsfont}
\usepackage[dvipsnames]{xcolor}
\usepackage[colorinlistoftodos]{todonotes}
\usepackage{booktabs}
\usepackage{xfrac}
\usepackage{bbm}

\usepackage{algorithm}
\usepackage{algorithmicx}

\usepackage[most]{tcolorbox}
\usepackage{xparse}
\usepackage{lipsum}
\usepackage{changepage}
\usepackage{enumitem}

\newcommand{\squishlist}{
   \begin{list}{$\bullet$}
    { \setlength{\itemsep}{0pt}      \setlength{\parsep}{3pt}
      \setlength{\topsep}{3pt}       \setlength{\partopsep}{0pt}
      \setlength{\leftmargin}{1.5em} \setlength{\labelwidth}{1em}
      \setlength{\labelsep}{0.5em} } }

\newcommand{\squishlisttwo}{
   \begin{list}{$\bullet$}
    { \setlength{\itemsep}{0pt}    \setlength{\parsep}{0pt}
      \setlength{\topsep}{0pt}     \setlength{\partopsep}{0pt}
      \setlength{\leftmargin}{2em} \setlength{\labelwidth}{1.5em}
      \setlength{\labelsep}{0.5em} } }

\newcommand{\squishend}{
    \end{list}  }

\newcommand{\two}{(2)}

\DeclareMathAlphabet{\mathpzc}{OT1}{pzc}{m}{n}

\usepackage[authoryear, sort&compress, round]{natbib} 
\usepackage{graphicx}
\usepackage{gensymb}
\usepackage{amsfonts}
\let\savedegree\degree
\let\degree\relax
\usepackage{mathabx}
\let\degree\savedegree

\usepackage{listings}
\usepackage{color}
\definecolor{deepblue}{rgb}{0,0,0.5}
\definecolor{deepred}{rgb}{0.6,0,0}
\definecolor{deepgreen}{rgb}{0,0.5,0}

\title{The StreetLearn Environment and Dataset}
\correspondingauthor{piotrmirowski@google.com}
\paperurl{}
\reportnumber{} 

\renewcommand{\today}{2019-03-04}

\author[*,1]{Piotr Mirowski}
\author[*,1]{Andras Banki-Horvath}
\author[*,1]{Keith Anderson}
\author[*,1]{Denis Teplyashin}
\author[2]{\\Karl Moritz Hermann}
\author[1]{Mateusz Malinowski}
\author[1]{Matthew Koichi Grimes}
\author[1]{Karen Simonyan}
\author[1]{\\Koray Kavukcuoglu}
\author[1]{Andrew Zisserman}
\author[1]{Raia Hadsell}

\affil[*]{Equal contributions}
\affil[1]{DeepMind, London, United Kingdom}
\affil[2]{DeepMind, Berlin, Germany}

\begin{abstract}
Navigation is a rich and well-grounded problem domain that drives progress in many different areas of research: perception, planning, memory, exploration, and optimisation in particular. Historically these challenges have been separately considered and solutions built that rely on stationary datasets---for example, recorded trajectories through an environment. These datasets cannot be used for decision-making and reinforcement learning, however, and in general the perspective of navigation as an interactive learning task, where the actions and behaviours of a learning agent are learned simultaneously with the perception and planning, is relatively unsupported.
 Thus, existing navigation benchmarks generally rely on static datasets~\citep{geiger2013vision,kendall2015posenet} or simulators~\citep{beattie2016deepmind,shah2018airsim}. To support and validate research in end-to-end navigation, we present StreetLearn: an interactive, first-person, partially-observed visual environment that uses Google Street View for its photographic content and broad coverage, and give performance baselines for a challenging goal-driven navigation task. The environment code, baseline agent code, and the dataset are available at \url{http://streetlearn.cc}. \end{abstract}

\begin{document}
\maketitle
\balance

\section{Introduction}
\label{sec:intro}
\begin{figure}[ht]
\begin{center}
\centerline{\includegraphics[width=\columnwidth]{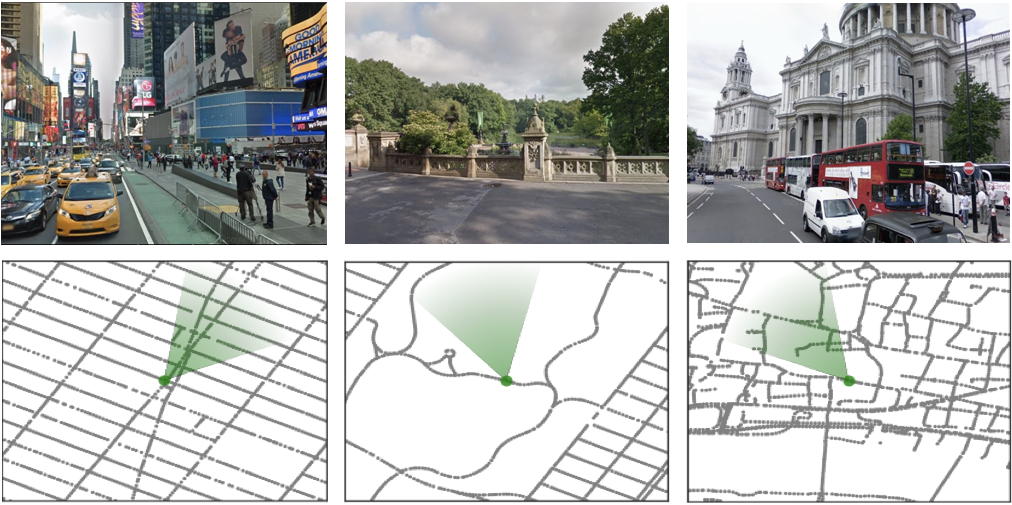}}
\caption{Our environment is built of real-world places from StreetView. The figure shows diverse views and corresponding local maps in New York City (Times Square, Central Park) and London (St.\ Paul's Cathedral). The green cone represents the agent's location and orientation.}
\label{fig:streetview}
\end{center}
\end{figure}

The subject of navigation is attractive to various research disciplines and technology domains alike, being at once a subject of inquiry from the point of view of neuroscientists wishing to crack the code of grid and place cells~\citep{banino2018vector,cueva2018emergence}, as well as a fundamental aspect of robotics research wishing to build mobile robots that can reach a given destination. The majority of navigation algorithms involve building an explicit map during an exploration phase and then planning and acting via that representation. 
More recently, researchers have sought to directly learn a navigation policy through exploration and interaction with the environment, for instance by using end-to-end deep reinforcement learning~\citep{zhu_icra2017, wu2018building, mirowski2016learning, lample_aaai17}. To support this research, we have designed an interactive environment called \emph{StreetLearn} that uses the images and underlying connectivity information from Google Street View (see Fig.~\ref{fig:streetview}) in two large areas comprising Pittsburgh and New York City. The environment features high-resolution photographic images displaying a diversity of urban settings, and spans city-scale areas with real-world street connectivity graphs.
Within this environment we have developed several traversal tasks that requires that the agent navigates from goal to goal over long distances. One such task has a real-world analogy of a \emph{courier} operating in a given city that starts at an arbitrary location called ``A'' and then is directed to go to a specific location ``B'' defined using absolute coordinates, without having ever been shown the map featuring these locations or the path going from A to B, or been told its own position. Another task consists in following step-by-step directions consisting of natural language navigation instructions and image thumbnails, mimicking Google Maps. Additional navigation tasks can be developed in the StreetLearn environment.

We describe the dataset, environment, and tasks in Section~\ref{sec:environment}, explain the environment code in Section~\ref{sec:code}, describe implemented approaches and baseline methods in Section~\ref{sec:methods} with results in Section~\ref{sec:results}, and detail related work in Section~\ref{sec:related}.

\section{Environment}
\label{sec:environment}
\begin{figure}[ht]
\begin{center}
\centerline{\includegraphics[width=.9\columnwidth]{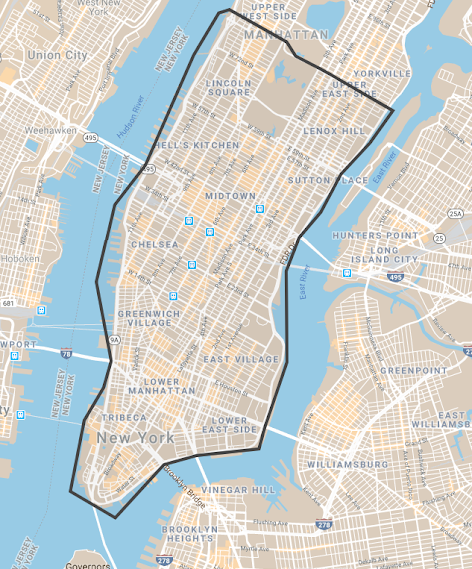}}
\vspace{.1in}
\centerline{\includegraphics[width=.9\columnwidth]{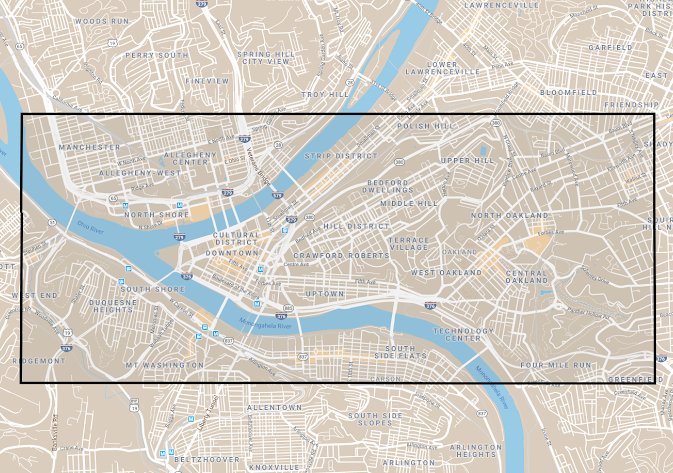}}
\caption{Maps with bounding boxes indicating the dataset coverage in New York City (top) and Pittsburgh (bottom).
}
\label{fig:bounds}
\end{center}
\end{figure}

\begin{figure}[ht]
\begin{center}
\centerline{\includegraphics[width=.9\columnwidth]{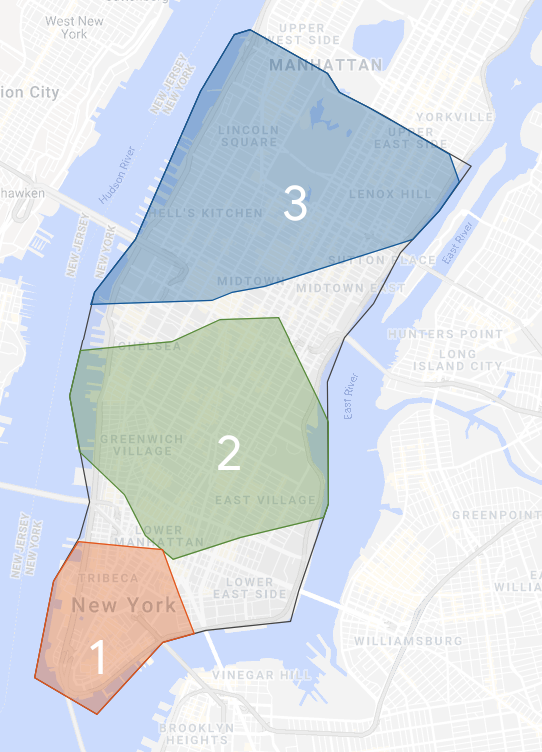}}
\vspace{.1in}
\centerline{\includegraphics[width=.9\columnwidth]{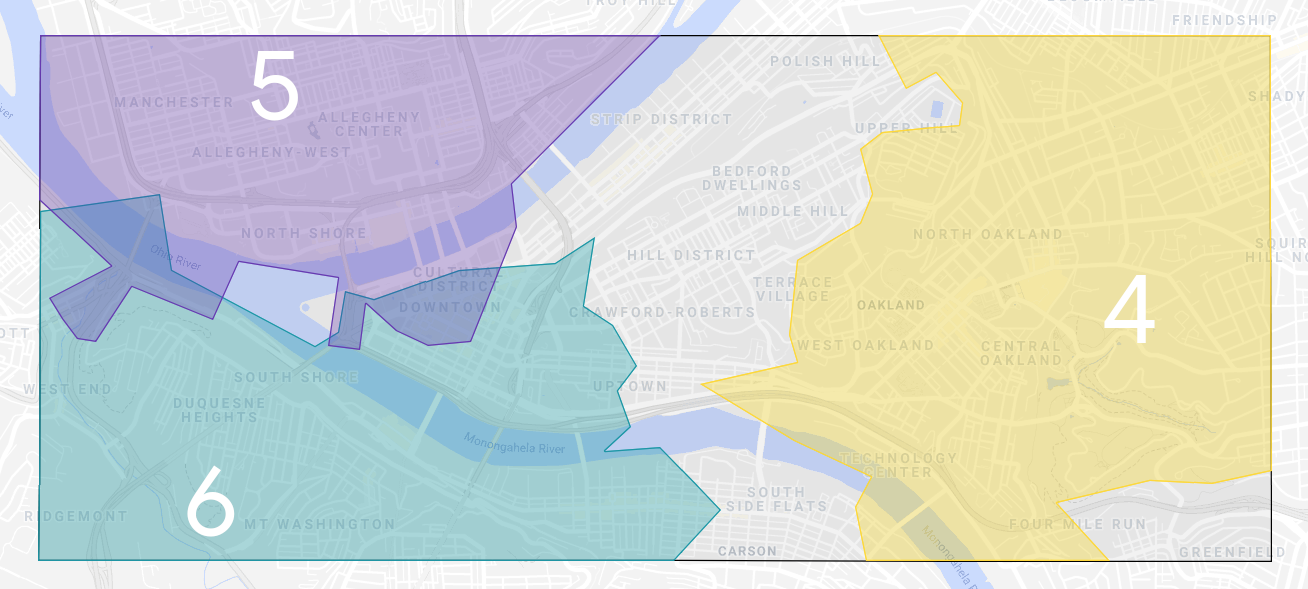}}
\caption{Maps with polygons delimiting the \emph{Wall Street} (1), \emph{Union Square} (2) and \emph{Hudson} (3) regions in New York City (top) and the \emph{CMU} (4), \emph{Allegheny} (5) and \emph{South Shore} (6) regions in Pittsburgh (bottom).
}
\label{fig:areas}
\end{center}
\end{figure}

This section presents StreetLearn, an interactive environment constructed using Google Street View. Since Street View data has been collected worldwide, and includes both high-resolution imagery and graph connectivity, it is a valuable resource for studying navigation (Fig.\ref{fig:streetview}). 

Street View provides a set of geolocated 360\degree\  panoramic images which form the nodes of an undirected graph (we use the term node and panorama interchangeably). We selected regions in New York City and Pittsburgh (see Fig.\ref{fig:bounds}). The area of New York City which is available for download is Manhattan south of 81st Street. This comprises approximates 56K panoramic images within a lat/long bounding box defined by $(40.695, -74.028)$ and $(40.788, -73.940)$. Note that Brooklyn, Queens, Roosevelt Island as well as the bridges and tunnels out of Manhattan are excluded, and we include only panoramas inside a polygon that follows the waterfront of Manhattan and 79th / 81st Street, covering an area of 31.6 km$^2$. The Pittsburgh dataset comprises 58K images and is defined by a lat/long bounding box between  $(40.425, -80.035)$ and $(40.460, -79.930)$, covering an area of 8.9 km by 3.9km. Additionally, we identify three regions in each city which can be used individually for training or for transfer learning experiments. The statistics of each region are given in Table~\ref{tab:regions}.

The undirected graph edges define the proximity and accessibility of nodes to other nodes. We do not reduce or simplify the underlying connectivity but rather use the full graph; thus there are congested areas with many nodes, complex occluded intersections, tunnels and footpaths, and other ephemera. The average node spacing is 10m, with higher densities at intersections. Although the graph is used to construct the environment, the agent never observes the underlying graph---only the RGB images are observed (overlay information, such as arrows, that are visible in the public Street View product are also not seen by the agent). Examples of the RGB images and the graph are shown in Figure~\ref{fig:streetview}.

In our dataset, each panorama is stored as a Protocol Buffer~\citep{protobuf2008} object, containing a string in high-quality compressed JPEG format that encodes the equirectangular image, and decorated with the following attributes: a unique string identifier, the position (lat/long coordinates and altitude in meters) and orientation (pitch, roll and yaw angles) of the panoramic camera, date of acquisition of the image, and a list of directly connected neighbours.

\subsection{Defining Areas Within the Dataset}
\label{sec:areas}

The whole Manhattan and Pittsburgh environments in the StreetLearn dataset encompass large urban areas that represent over 56k Street View panoramas each, and traversing these areas from one extremity to another could entail going through close to 1k nodes in the Street View graph. To make learning tractable and also to define distinct regions for training and transfer, one can cut the environment into smaller areas. For instance, Figure~\ref{fig:areas} illustrates a cut of Manhattan and Pittsburgh into 6 regions ("Wall Street", "Union Square", "Hudson", "CMU", "Allegheny" and "South Shore") that used in our experiments in Section~\ref{sec:results}. 

There are many possibilities to define areas inside a street graph: the most obvious is to cut the graph using a latitude/longitude bounding box, with the disadvantage of creating unconnected components. The second is to cut the graph using a polygon, with the inconvenience of having to specify all the vertices of that polygon, relying on convex hulls to select the nodes included within the polygon. We chose a third approach for defining our areas, by growing graph areas by Breadth-First Search (BFS)\citep{zuse1972plankalkul,moore1959shortest} from a given node, which requires to choose only a central panorama and a graph depth, and which ensure that the resulting graph is connected. We list in Table~\ref{tab:regions} the size (in nodes, edges and area coverage), the elevation changes and a description of those areas, including the central panorama ID and the BFS graph depth.

\subsection{Agent Interface and the Courier Task}
\label{sec:agent}

An RL environment needs to specify the observations and action space of the agent as well as define the task. The StreetLearn environment provides a visual observation at each timestep,  ${\bf x_t}$. The visual inputs are meant to simulate a first-person, partially observed environment, thus ${\bf x_t}$ is a cropped, $60\degree$ square, RGB image that is scaled to $84\times84$ pixels (i.e.\ not the entire panorama).

The action space is composed of five discrete actions: ``slow" rotate left or right ($\pm 22.5\degree$), ``fast" rotate left or right ($\pm 67.5\degree$), or move forward (this action becomes a \emph{noop} if there is not an edge in view from the current agent pose). If there are multiple edges in the viewing cone of the agent, then the most central one is chosen. 

StreetLearn provides an additional observation, the goal descriptor $g_t$, which communicates the task objective to the agent---where to go to receive the next reward. There are many options for how to specify the goal: e.g., images are a natural choice (as in~\citep{zhu_icra2017}) but quickly become ambiguous at city scale; language-based directions or street addresses could be used (as in~\citep{chen2018touchdown}) though this would place the emphasis on language grounding rather than navigation; and landmarks could be used to encode the target location in a scalable, coordinate-free way~\citep{mirowski2018learningcityscale}. For this \emph{courier} task we take the simplest route and define goal locations straightforwardly as continuous-valued coordinates $(Lat^g_t,Long^g_t)$. Note that the goal description is \emph{absolute}; it is not relative to the agent's position and only changes when a new goal is drawn (either upon successful goal acquisition or at the beginning of an episode).

In the \emph{courier} task, which can be summarised as the problem of navigating to a series of random locations in a city, the agent starts each episode from a randomly sampled position and orientation within the StreetLearn graph. A goal location is randomly sampled from the graph and the goal descriptor $g_0$ is computed and input to the agent. If the agent reaches a node that is near to the goal (100m, or approximately one city block), the agent is rewarded and the next goal is randomly chosen and input to the agent. 
Each episode ends after 1000 agent steps.  The reward that the agent gets upon reaching a goal is proportional to the shortest path between the goal and the agent's position when the goal is first assigned; much like a delivery service, the agent receives a higher reward for longer journeys.  

Intuitively, in order to solve the courier task, the agent will need to learn to associate the goal encoding with the images observed at the goal location, as well as to associate the images observed at the current location with the policy to reach different goal locations.

\begin{table*}[ht]
    \scriptsize
    \setlength\tabcolsep{7pt} 
    \centering
    \begin{tabular}{@{}lcccccp{6cm}@{}}
    \toprule
    Region & \#nodes & \#edges & av. edge len. & elev. change & area & description \\
    \midrule
    Wall Street & 7224 & 7496 & 9.8m & 31m & 3.8km$^2$ & Southernmost area of Manhattan, skyscrapers, narrow streets and highways with irregular intersections. Graph of depth 215, centered at pano {\tt 6rIMyvAZUW4sT3ffqYOg0w}. \\
    \midrule
    Union Square & 15525 & 16094 & 9.8m & 40m & 9.7km$^2$ & Between Downtown and Midtown Manhattan; skyscrapers, brownstones and townhouses; parks and regular street grid. Graph of depth 200, centered at pano {\tt dFbip4uo7CNu86y52Axc5g}. \\
    \midrule
    Hudson River & 18085 & 18676 & 9.9m & 56m & 11.7km$^2$ & Riverside along Hudson River and near Central Park; skyscrapers, regular street grid and highways. Graph of depth 400, centered at pano {\tt PreXwwylmG23hnheZ\_\_zGw}. \\
    \midrule
    CMU & 15947 & 16339 & 9.9m & 146m & 11.2km$^2$ & Suburban areas of Oakland near CMU, suburban, leafy streets with high altitude differentials. Graph of depth 400, centered at pano {\tt r5DqC1wcUi2Lw6T4GvUxwQ}. \\
    \midrule
    Allegheny & 14073 & 14567 & 9.8m & 104m & 7.2km$^2$ & Downtown Pittsburgh and historic district, large avenues, highways and bridges. Graph of depth 320, centered at pano {\tt ohwj1wXoJ3KOPwnSPaAMCw}. \\
    \midrule
    South Shore & 14967 & 15370 & 9.9m & 151m & 9.4km$^2$ & Downtown Pittsburgh, South Shore, South Side Flats and Duquesne Heights, highways, bridges and long tunnels, funicular. Graph of depth 350, centered at pano {\tt ljBFHUcoonDeE2omJ7PrOQ}. \\
    \bottomrule
    \end{tabular}
    \caption{Relevant information for the three regions in New York (Wall Street, Union Square, and Hudson River) and three regions in Pittsburgh (CMU, Allegheny, and South Shore).}
    \label{tab:regions}
\end{table*}

\subsection{Curriculum}
\label{sec:curriculum}

Curriculum learning gradually increases the complexity of the learning task by choosing more and more difficult examples to present to the learning algorithm 
\citep{bengio2009curriculum, graves2017automated,zaremba2014learning}. We have found that a curriculum may be important for the courier task with more distant destinations. Similar to other RL problems such as Montezuma's Revenge, the courier task suffers from very sparse rewards; unlike that game, we are able to define a natural curriculum scheme. We start by sampling new goals within 500m of the agent's position (phase 1).
In phase 2, we progressively grow the maximum range of allowed goals to cover the full graph.

Note that while this paper focuses on the \emph{courier} task, but as described in the following Section~\ref{sec:code}, the environment has been enriched with the possibility of specifying directions through step-by-step pairs of (image, natural language instruction) and goal image.

\section{Code}
\label{sec:code}
\subsection{Code Structure}
\label{sec:structure}

We have made the environment and agent code available at \url{https://github.com/deepmind/streetlearn}. The code repository contains the following components:

\begin{itemize}
\item Our C++ StreetLearn engine for loading, caching and serving Google Street View panoramas as well as for handling navigation (moving from one panorama to another) depending on the city street graph and the current position and orientation of the agent. Each panorama is projected from its equirectangular~\citep{equirectangular2005} representations to a first-person view for which one can specify the yaw, pitch and field of view angles.
\item The message protocol buffers~\citep{protobuf2008} used to store the panoramas and the street graph.
\item A Python-based interface for calling the StreetLearn environment with custom action spaces.
\item Within the Python StreetLearn interface, several games are defined in individual files whose names end with game.py.
\item A simple human agent, implemented in Python using Pygame\footnote{\url{https://www.pygame.org}}, that instantiates the StreetLearn environment on the requested map and enables a user to play the courier or the instruction-following games.
\item Oracle agents, similar to the human agent, which automatically navigate towards a specified goal and reports oracle performance on the courier or instruction-following games.
\item TensorFlow implementation of agents.
\end{itemize}

\subsection{Code Interface}
\label{sec:interface}

Our Python StreetLearn environment follows the specifications from OpenAI Gym\footnote{\url{https://gym.openai.com/}}~\citep{brockman2016openai}.

After instantiating a specific game and the environment, the environment can be initialised by calling function {\tt {\bf reset}}$()$. Note that if the flag {\tt {\bf auto\_reset}} is set to \emph{True} at construction, {\tt {\bf reset}}$()$ will be called automatically every time that an episode ends.

As illustrated in Listing \ref{code:loop}, the agent plays within the environment by iteratively producing an action, sending it to (\emph{stepping} through) the environment, and processing the observations and rewards returned by the environment. The call to function {\tt {\bf step(action)}} returns:

\begin{itemize}
    \item {\tt {\bf observation}} (tuple of observations arrays and scalars that are requested at construction),
    \item {\tt {\bf reward}} (a floating-point scalar number with the current reward of the agent),
    \item {\tt {\bf done}} (boolean indicating whether a game episode has ended and been reset),
    \item and {\tt {\bf info}} (a dictionary of environment state variables, which is useful for debugging the agent behaviour or for accessing privileged environment information for visualisation and analysis).
\end{itemize}

\subsection{Actions and observations}
\label{sec:actions}

We have made four actions available to the agent:

\begin{itemize}
    \item Rotate left or right in the panorama, by a specified angle (change the yaw of the agent).
    \item Rotate up or down in the panorama, by a specified angle (change the pitch of the agent).
    \item Move from current panorama A forward to another panorama B if the current bearing of the agent from A to B is within a tolerance angle of 30 degrees.
    \item Zoom in and out in the panorama.
\end{itemize}

As such, agent actions are sent to the environment via {\tt {\bf step(action)}} as tuples of 4 scalar numbers. However, for training discrete policy agents via reinforcement learning, action spaces are discretised into integers. For instance, we used 5 actions in \citep{mirowski2018learningcityscale}: (move forward, turn left by 22.5 deg, turn left by 67.5 deg, turn right by 22.5 deg, turn right by 67.5 deg).

The following observations can currently be requested from the environment:

\begin{itemize}
    \item {\tt {\bf view\_image}}: RGB image for the first-person view image returned from the environment and seen by the agent,
    \item {\tt {\bf graph\_image}}: RGB image for the top-down street graph image, usually not seen by the agent,
    \item {\tt {\bf pitch}}: Scalar value of the pitch angle of the agent, in degrees (zero corresponds to horizontal),
    \item {\tt {\bf yaw}}: Scalar value of the yaw angle of the agent, in degrees (zero corresponds to North),
    \item {\tt {\bf yaw\_label}}: Integer discretized value of the agent yaw using 16 bins,
    \item {\tt {\bf metadata}}: Message protocol buffer of type Pano with the metadata of the current panorama,
    \item {\tt {\bf target\_metadata}}: Message protocol buffer of type Pano with the metadata of the target/goal panorama,
    \item {\tt {\bf latlng}}: Tuple of lat/lng scalar values for the current position of the agent,
    \item {\tt {\bf latlng}}: Integer discretized value of the current agent position using 1024 bins (32 bins for latitude and 32 bins for longitude),
    \item {\tt {\bf target\_latlng}}: Tuple of lat/lng scalar values for the target/goal position,
    \item {\tt {\bf target\_latlng}}: Integer discretized value of the target position using 1024 bins (32 bins for latitude and 32 bins for longitude),
    \item {\tt {\bf thumbnails}}: set of $n+1$ RGB images for the first-person view image returned from the environment, that should be seen by the agent at specific waypoints and goal locations when playing the instruction-following game with $n$ instructions,
    \item {\tt {\bf instructions}}: set of $n$ instructions for the agent at specific waypoints and goal locations when playing the instruction-following game with $n$ instructions,
    \item {\tt {\bf neighbors}}: Vector of immediate neighbor egocentric traversability grid around the agent, with 16 bins for the directions around the agent and bin 0 corresponding to the traversability straight ahead of the agent.
    \item {\tt {\bf ground\_truth\_direction}}: Scalar value of the relative ground truth direction to be taken by the agent in order to follow a shortest path to the next goal or waypoint. This observation should be requested only for agents trained using imitation learning.
\end{itemize}

\subsection{Games}
\label{sec:games}

The following games are available in the StreetLearn environment:

\subsubsection{coin\_game}
In the \emph{coin\_game}, the rewards consist in invisible coins scattered throughout the map, yielding a reward of 1 for each. Once picked up, these rewards do not reappear until the end of the episode.

\subsubsection{courier\_game}
In the \emph{courier\_game}, the agent is given a goal destination, specified as lat/long pairs. Once the goal is reached (with 100m tolerance), a new goal is sampled, until the end of the episode. Rewards at a goal are proportional to the number of panoramas on the shortest path from the agent's position when it gets the new goal assignment to that goal position. Additional reward shaping consists in early rewards when the agent gets within a range of 200m of the goal. Additional coins can also be scattered throughout the environment. The proportion of coins, the goal radius and the early reward radius are parameterizable. The \emph{curriculum\_courier\_game} is similar to the \emph{courier\_game}, but with a curriculum on the difficulty of the task (maximum straight-line distance from the agent's position to the goal when it is assigned).

\subsubsection{Instruction games}
The \emph{goal\_instruction\_game} and its variations \emph{incremental\_instruction\_game} and \emph{step\_by\_step\_instruction\_game} use navigation instructions to direct agents to a goal. Agents are provided with a list of instructions as well as thumbnails that guide the agent from its starting position to the goal location. In step\_by\_step, agents are provided one instruction and two thumbnails at a time, in the other game variants the whole list is available throughout the whole game. Reward is granted upon reaching the goal location (all variants), as well as when hitting individual waypoints (\emph{incremental} and \emph{step\_by\_step} only). During training various curriculum strategies are available to the agents, and reward shaping can be employed to provide fractional rewards when the agent gets within a range of 50m of a waypoint or goal.

\begin{figure*}
\small
\begin{lstlisting}[language=Python,frame=tb,keywordstyle=\bf\color{deepred},emph={game,courier_game,streetlearn,config,env,action,done,info,reward,sum_rewards,observation},emphstyle=\bf\color{deepblue}]
# Instantiate a game (each game has its own class and constructor).
game = courier_game.CourierGame(config)
env = streetlearn.StreetLearn(FLAGS.dataset_path, config, game)
env.reset()
action = np.array([0, 0, 0, 0])
sum_rewards = 0
while True:
    observation, reward, done, info = env.step(action)
    # Plot the observations.
    # [...]
    # Keep track of episode ends and of rewards.
    sum_rewards += reward
    if done:
        sum_rewards = 0
    # Use info for analysing the agent performance on the game.
    # [...]
    # Take an action
    action = some_agent_function(observation)
\end{lstlisting}
\caption{Main loop for interacting with the environment.}
\label{code:loop}
\end{figure*}

\section{Methods}
\label{sec:methods}
This section briefly describes the set of approaches which are evaluated on the courier task.

\subsection{Goal-dependent Actor-Critic Reinforcement Learning}
\label{sec:problem}
We formalise the learning problem as a Markov Decision Process, with
state space $\mathcal{S}$, action space $\mathcal{A}$, environment $\mathcal{E}$, and a set of possible goals $\mathcal{G}$.

The reward function depends on  the current goal and state: $R: \mathcal{S} \bigtimes \mathcal{G} \bigtimes \mathcal{A} \to \mathbb{R}$. 
The usual reinforcement learning objective is to find the policy that maximises the expected return defined as the sum of discounted rewards starting from state $s_0$ with discount $\gamma$. 
In this navigation task, the expected return from a state $s_t$ also depends on the series of sampled goals $\{g_{k}\}_k$. The policy is a distribution over actions given the current state $s_t$ and the goal $g_t$: $\pi(a|s,g) = Pr(a_t=a |s_t=s, g_t=g)$. We define the value function to be the expected return for the agent that is sampling actions from policy $\pi$ from state $s_t$ with goal $g_t$:  $V^{\pi}(s,g)=E[R_t]= E[\, \sum_{k=0}^{\infty} \gamma^k r_{t+k} |s_t=s,g_t=g]$.

\begin{figure}[ht]
\begin{center}
\centerline{\includegraphics[width=\columnwidth]{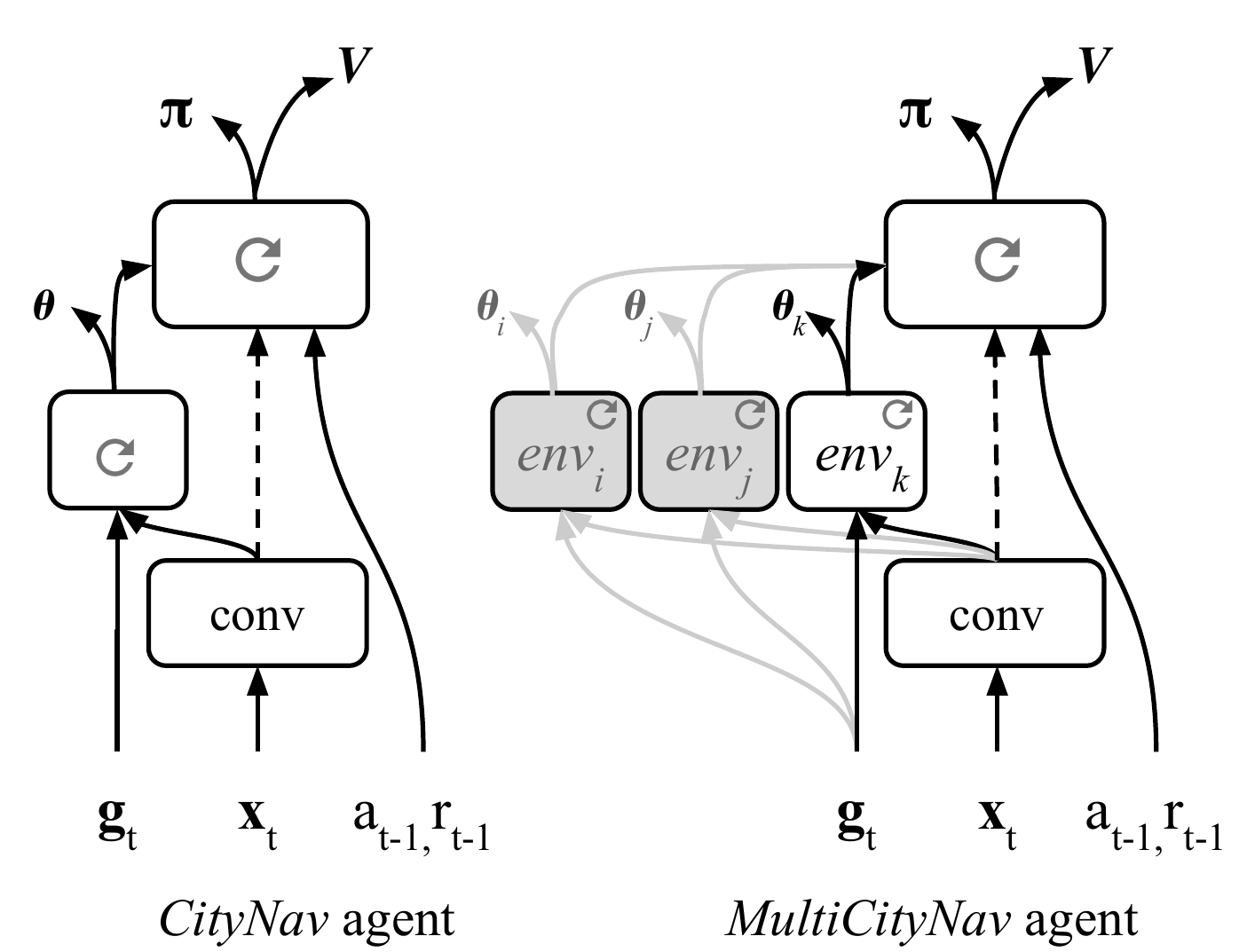}}
\caption{Comparison of architectures. \emph{Left:} CityNav is a single-city navigation architecture with a policy LSTM, a separate goal LSTM, and optional auxiliary heading ($\theta$). \emph{Right:} MultiCityNav is a multi-city architecture with individual goal LSTM pathways for each city.
}
\label{fig:architecture}
\end{center}
\vskip -0.2in
\end{figure}

We hypothesise that an agent should benefit from two types of learning: first, learning a general and location-agnostic representation and exploration behaviour, and second, learning locale-specific structure and features. A navigating agent not only needs an internal representation that is general, to support cognitive processes such as scene understanding, but also needs to organise and remember the features and structures that are unique to a place. Therefore, to support both types of learning,
we focus on  neural architectures with multiple pathways. 

We evaluate two agents on the six regions described in Table~\ref{tab:regions}. We give here a summary of the approach, as the full architectural details of these agents have been previously described~\citep{mirowski2018learningcityscale}.
The policy and the value function are both parameterised by a neural network which shares all layers except the final linear outputs. The agent operates on raw pixel images ${\bf x_t}$, which are passed through a convolutional network as in \citep{mnih2016asynchronous}. A Long Short-Term Memory (LSTM) \citep{hochreiter1997long} receives the output of the convolutional encoder as well as the past reward $r_{t-1}$ and previous action $a_{t-1}$.  The two different architectures are described below. 

The {\bf CityNav} architecture (Fig.~\ref{fig:architecture}b) has a convolutional encoder and two LSTM layers, which are designated as a \emph{policy LSTM} and a \emph{goal LSTM}. The goal description $g_t$ is input to the goal LSTM along with the previous action and reward, as well as the visual features from the convolutional encoder. The CityNav agent also adds an auxiliary heading ($\theta$)  prediction task on the outputs of the goal LSTM.
    
The {\bf MultiCityNav} architecture (Fig.~\ref{fig:architecture}c) extends the CityNav agent to learn in different cities. The remit of the goal LSTM is to encode and encapsulate locale-specific features and topology such that multiple pathways may be added, one per city or region. Moreover, after training on a number of cities, we demonstrate that the convolutional encoder and the policy LSTM become general enough that only a new goal LSTM needs to be trained for new cities. 

To train the agents, we use IMPALA \citep{espeholt2018impala}, an actor-critic implementation that decouples acting and learning. In our experiments, IMPALA results in similar performance to A3C \citep{mnih2016asynchronous}. We use 256 actors for \emph{CityNav} and 512 actors for \emph{MultiCityNav}, with batch sizes of 256 or 512 respectively, and sequences are unrolled to length 50.

We note that these computational resources are not available for all, so we have verified that comparable results are attained using only 16 actors and 1 learner, running on a single desktop computer with a Graphics Processing Unit (GPU). The desktop we used had large memory (192 GB) for instantiating 16 StreetLearn environments (each environment requiring a large cache memory for caching panoramas), but smaller memory could be used as well with the trade-off of more frequent disk accesses.

A TensorFlow implementation of the \emph{CityNav} and baseline architectures from \citep{mirowski2018learningcityscale} is made available on the code repository at \url{https://github.com/deepmind/streetlearn}. The trainer code is a directy modification of \citep{espeholt2018impala} from \url{https://github.com/deepmind/scalable_agent} and is made available separately.

\subsection{Oracle}
\label{sec:oracle}

We also compute an upper bound for all the tasks by computing the shortest path from all panorama positions to the specified goal position using breadth-first search~\citep{zuse1972plankalkul,moore1959shortest} on the panorama connectivity graph. This enables us to calculate both which is the next panorama that agent should go to and the direction that the agent should align with in order to move forward to that panorama, repeating this process until arriving at destination. This \emph{ground\_truth\_position} can be requested as an observation (for imitation learning agents) or be taken from the {\tt {\bf info}} dictionary returned by the environment. Listing \ref{code:oracle} shows how the \emph{oracle} agent can be implemented to provide with a valuable measure to benchmark the tasks.

\begin{figure*}
\small
\begin{lstlisting}[language=Python,frame=tb,keywordstyle=\bf\color{deepred},emph={game,courier_game,streetlearn,config,env,action,action_spec,bearing,done,info,reward,sum_rewards,observation},emphstyle=\bf\color{deepblue},stringstyle=\bf\color{deepgreen}]
game = courier_game.CourierGame(config)
env = streetlearn.StreetLearn(FLAGS.dataset_path, config, game)
env.reset()
action = np.array([0, 0, 0, 0])
action_spec = env.action_spec()
while True:
    observation, reward, done, info = env.step(action)
    # Plot the observations.
    # [...]
    bearing = info["bearing_to_next_pano"]
    if bearing > FLAGS.horizontal_rot:
        action = FLAGS.horizontal_rot * action_spec["horizontal_rotation"]
    elif bearing < -FLAGS.horizontal_rot:
        action = -FLAGS.horizontal_rot * action_spec["horizontal_rotation"]
    else:
        action = action_spec["move_forward"]
\end{lstlisting}
\caption{Oracle implementation using the ground truth direction/bearing to the next panorama.}
\label{code:oracle}
\end{figure*}

\section{Results on the Courier Task}
\label{sec:results}
To evaluate the described approaches, we give the individual performance in each region as well as the result of training jointly over multiple regions. We also show the capability of the approach to generalise by evaluating goals in held-out areas, and by training only part of the agent for an entirely new region. 

Table~\ref{tab:city} gives the average total reward per 1000-step episode achieved by different agents in six different regions of New York City and Pittsburgh defined on Figure~\ref{fig:areas} and Table~\ref{tab:regions}. Although the agents were trained with reward shaping (i.e., they receive partial rewards when they are within a small radius of the goal), the per-episode returns given here only include the full reward which is given when the goal is reached. The experiments are all replicated with 5 different seeds.

In Table~\ref{tab:city}, \emph{Oracle} results are the result of breadth-first search directly on the graph; hence they reflect perfect performance. \emph{Single} results show the performance of agents trained individually for each region using the \emph{CityNav} architecture. The trained agents do well in New York City, achieving 85\% to 97\% of oracle returns, and do less well in Pittsburgh, particularly in the South Shore region where the agent fails completely. This is presumably due to the challenging elevation changes in the region which give rise to convoluted routes even between nearby nodes, and is an artifact of how we specify the curriculum task (based on the maximum Euclidean distance from the agent position to the goal, not accounting for actual travel time). Specifically, when the agent is at the top of Duquesne Hill in South Shore, a goal location on the other side of the river and that is 500m away by bird flight might be kilometres away by road distance. 

\emph{Joint} results show the per-region performance of a \emph{MultiCityNav} agent that is trained jointly across five regions (South Shore is excluded). The resulting agent suffers only a small drop in performance even though it is now trained across a much broader area: two cities and five regions. Finally, \emph{transfer} gives the performance of an agent that is trained on four regions (given in italics) and then transferred to a fifth region (Wall Street). In this transfer, only the goal LSTM is modified; there are no gradient updates to the other two components of the architecture (the convolutional encoder or the policy LSTM). 

\begin{table}[ht]
    \scriptsize
    \setlength\tabcolsep{6pt} 
    \centering
    \begin{tabular}{@{}lcccc@{}}
    \toprule
    City & Oracle & Single & Joint & Transfer \\
    \midrule
    Wall Street  & 809 & 782 & 745 & 541 \\
    Union Square & 750 & 721 & 681 & \emph{667} \\
    Hudson River & 721 & 615 & 621 & \emph{601} \\
    \midrule
    CMU  & 755 & 473 & 313 & \emph{355} \\
    Allegheny & 760 & 669 & 571 & \emph{562} \\
    South Shore & 737 & 1 & - & - \\
    \bottomrule
    \end{tabular}
    \caption{Per-city goal rewards for Oracle, single-trained \emph{CityNav} as well as \emph{MultiCityNav} agents trained jointly on 5 cities (Wall Street, Union Square and Hudson River in Manhattan, CMU and Allegheny in Pittsburgh) or jointly on 4 cities (Union Square, Hudson River, CMU and Allegheny) then transferred to Wall Street.}
    \label{tab:city}
\end{table}

To investigate the generalisation capability of a trained agent, we mask $25\%$ of the possible goals and train on the remaining ones (see Figure 5 in \citep{mirowski2018learningcityscale} for an illustration). At test time we evaluate the agent only on its ability to reach goals in the held-out areas. Note that the agent is still able to traverse \emph{through} these areas, it just never samples a goal there.
More precisely, the held-out areas are squares
sized $0.01\degree$ (coarse grid) or $0.005\degree$ (medium grid) of latitude and longitude (respectively roughly about 1km$^2$ and 0.5km$^2$). 

In the experiments, we train the \emph{CityNav} agent for 1B steps, and next freeze the weights of the agent and evaluate its performance on held-out areas for 100M steps. Table \ref{tab:heldout} shows some decreasing performance of the agents as the held-out area size increases. 
To gain further understanding, in addition to \emph{Test Reward} metric, we also use missed goals (\emph{Fail}) and half-trip time ($T_{\frac{1}{2}}$) metrics. The missed goals metric measures the percentage of times goals were not reached. The half-trip time measures the number of agent steps necessary to cover half the distance separating the agent from the goal.

\begin{table}[ht]
    \scriptsize
    \setlength\tabcolsep{6pt} 
    \centering
    \begin{tabular}{@{}lcccc@{}}
    \toprule
    Grid size & Area & Goal rewards & Fail & $T\frac{1}{2}$ \\
    \midrule
    No grid & Train & 719 & 0\% & 133  \\
    \midrule
    Medium grid & Held-out & 724 & 0\% & 126 \\
    Coarse grid & Held-out & 605 & 2\% & 164 \\
    \bottomrule
    \end{tabular}
    \caption{\emph{CityNav} agent generalisation performance (reward and fail metrics) on a set of
held-out goal locations (medium and coarse grids). We also compute the half-trip time ($T\frac{1}{2}$), to reach halfway to the goal.}
    \label{tab:heldout}
\end{table}

We also compare, in Table~\ref{tab:target}, the performance achieved when using (lat, long) goal descriptors versus the previously proposed landmark descriptors~\citep{mirowski2018learningcityscale}. Although the landmark scheme has advantages, such as avoiding fixed coordinate frames, the (lat, long) descriptor is shown to out-perform landmarks on the Union Square region in New York.

\begin{table}[ht]
    \scriptsize
    \setlength\tabcolsep{6pt} 
    \centering
    \begin{tabular}{@{}lc@{}}
    \toprule
    Target representation & Goal rewards \\
    \midrule
    Oracle & 750 \\
    \midrule
    (lat, long) scalars & 721 \\ 
    Landmarks & 700 \\ 
    \bottomrule
    \end{tabular}
    \caption{\emph{CityNav} agent performance on Union Square with different types of target representations: (lat, long) scalars vs. landmarks.}
    \label{tab:target}
\end{table}

\section{Related Work}
\label{sec:related}
The StreetLearn environment is related to a number of other simulators and datasets that have emerged in recent years in response to a greater interest in reinforcement learning and, more generally, learning navigation through interaction. We focus on enumerating these related datasets and environments, referring the reader to~\citet{mirowski2018learningcityscale} for a more complete discussion of related approaches.

Many RL-based approaches for navigation rely on simulators which have the benefit of features like procedurally generated variations but tend to be visually simple and unrealistic, including synthetic 3D environments such as VizDoom~\citep{kempka2016vizdoom}, DeepMind Lab~\citep{beattie2016deepmind}, HoME~\citep{brodeur2017home}, House 3D~\citep{wu2018building}, Chalet~\citep{yan2018chalet}, or AI2-THOR~\citep{kolve2017ai2}.

To bridge the gap between simulation and reality, researchers have developed more realistic, higher-fidelity simulated environments \citep{dosovitskiy2017carla,kolve2017ai2,shah2018airsim,wu2018building}. However, in spite of their increasing photo-realism, the inherent problems of simulated environments lie in the limited diversity of the environments and the antiseptic cleanliness of the observations. Our real-world dataset is diverse and visually realistic, comprising scenes with pedestrians, cars, buses or trucks, diverse weather conditions and vegetation and covering large geographic areas. However, we note that there are obvious limitations of our environment: It does not contain dynamic elements, the action space is necessarily discrete as it must jump between panoramas, and the street topology cannot be arbitrarily altered or regenerated.

More visually realistic environments such as Matterport Room-to-Room~\citep{chang2017matterport3d}, AdobeIndoorNav~\citep{mo2018adobeindoornav}, Stanford 2D-3D-S~\citep{armeni_cvpr16}, ScanNet~\citep{dai2017scannet}, Gibson Env~\citep{xia2018gibson}, and MINOS~\citep{savva2017minos} have been recently introduced to represent indoor scenes, some augmented with navigational instructions.

Using New York imagery, \citet{de2018talk} use navigation instructions but rely on categorical annotation of nearby landmarks rather than visual observations and use a dataset of 500 panoramas only (ours is two orders of magnitude larger).
Very recently, \citet{cirikfollowing} and particularly \citet{chen2018touchdown} have also proposed larger datasets of driving instructions grounded in Street View imagery. 

\section{Conclusion}
\label{sec:conclusion}
Navigation is an important cognitive task that enables humans and animals to traverse a complex world without maps. To help understand this cognitive skill, its emergence and robustness, and its application to real-world settings, we have made public a dataset and an interactive environment based on Google Street View. Our carefully curated dataset has been constituted from photographic images that have been manually reviewed and vetted for privacy - we took these extra precautions to ensure that all faces and license plates are blurred appropriately. The dataset is made available at \url{http://streetlearn.cc} and is distributed on request; in the case when an individual requests a specific panorama to be taken down or to be blurred on the Google Street View website, we propagate their request to the users of the StreetLearn dataset and provide users with an updated version that complies with the takedown request.

Our environment enables the training of agents to navigate to different goal locations based purely on visual observations and absolute target position representations. We have also expanded that dataset with text instructions to enable reward-based task focused on following relative directions to reach a goal. We will rely on this dataset and environment to address the fundamental problem of grounded, long-range, goal-driven navigation.

\section*{Acknowledgements}

The authors wish to acknowledge Lasse Espeholt and Hubert Soyer for technical help with the IMPALA algorithm, Razvan Pascanu, Ross Goroshin, Phil Blunsom, and Nando de Freitas for their feedback, Chloe Hillier, Razia Ahamed, Richard Ives and Vishal Maini for help with the project, and the Google Maps and Google Street View teams for their support in accessing the data.

\bibliography{streetlearn}

\begin{thebibliography}{36}
\providecommand{\natexlab}[1]{#1}
\providecommand{\url}[1]{\texttt{#1}}
\expandafter\ifx\csname urlstyle\endcsname\relax
  \providecommand{\doi}[1]{doi: #1}\else
  \providecommand{\doi}{doi: \begingroup \urlstyle{rm}\Url}\fi

\bibitem[Armeni et~al.(2016)Armeni, Sener, Zamir, Jiang, Brilakis, Fischer, and
  Savarese]{armeni_cvpr16}
Iro Armeni, Ozan Sener, Amir~R. Zamir, Helen Jiang, Ioannis Brilakis, Martin
  Fischer, and Silvio Savarese.
\newblock 3d semantic parsing of large-scale indoor spaces.
\newblock In \emph{Proceedings of the IEEE International Conference on Computer
  Vision and Pattern Recognition (CVPR)}, 2016.

\bibitem[Banino et~al.(2018)Banino, Barry, Uria, Blundell, Lillicrap, Mirowski,
  Pritzel, Chadwick, Degris, Modayil, Wayne, Soyer, Viola, Zhang, Goroshin,
  Rabinowitz, Pascanu, Beattie, Petersen, Sadik, Gaffney, King, Kavukcuoglu,
  Hassabis, Hadsell, and Kumaran]{banino2018vector}
Andrea Banino, Caswell Barry, Benigno Uria, Charles Blundell, Timothy
  Lillicrap, Piotr Mirowski, Alexander Pritzel, Martin~J Chadwick, Thomas
  Degris, Joseph Modayil, Greg Wayne, Hubert Soyer, Fabio Viola, Brian Zhang,
  Ross Goroshin, Neil Rabinowitz, Razvan Pascanu, Charlie Beattie, Stig
  Petersen, Amir Sadik, Stephen Gaffney, Helen King, Koray Kavukcuoglu, Demis
  Hassabis, Raia Hadsell, and Dharshan Kumaran.
\newblock Vector-based navigation using grid-like representations in artificial
  agents.
\newblock \emph{Nature}, 557\penalty0 (7705):\penalty0 429, 2018.

\bibitem[Beattie et~al.(2016)Beattie, Leibo, Teplyashin, Ward, Wainwright,
  K{\"u}ttler, Lefrancq, Green, Vald{\'e}s, Sadik, et~al.]{beattie2016deepmind}
Charles Beattie, Joel~Z Leibo, Denis Teplyashin, Tom Ward, Marcus Wainwright,
  Heinrich K{\"u}ttler, Andrew Lefrancq, Simon Green, V{\'\i}ctor Vald{\'e}s,
  Amir Sadik, et~al.
\newblock Deepmind lab.
\newblock \emph{arXiv preprint arXiv:1612.03801}, 2016.

\bibitem[Bengio et~al.(2009)Bengio, Louradour, Collobert, and
  Weston]{bengio2009curriculum}
Yoshua Bengio, J{\'e}r{\^o}me Louradour, Ronan Collobert, and Jason Weston.
\newblock Curriculum learning.
\newblock In \emph{Proceedings of the 26th annual international conference on
  machine learning}, pages 41--48. ACM, 2009.

\bibitem[Brockman et~al.(2016)Brockman, Cheung, Pettersson, Schneider,
  Schulman, Tang, and Zaremba]{brockman2016openai}
Greg Brockman, Vicki Cheung, Ludwig Pettersson, Jonas Schneider, John Schulman,
  Jie Tang, and Wojciech Zaremba.
\newblock Openai gym.
\newblock \emph{arXiv preprint arXiv:1606.01540}, 2016.

\bibitem[Brodeur et~al.(2017)Brodeur, Perez, Anand, Golemo, Celotti, Strub,
  Rouat, Larochelle, and Courville]{brodeur2017home}
Simon Brodeur, Ethan Perez, Ankesh Anand, Florian Golemo, Luca Celotti, Florian
  Strub, Jean Rouat, Hugo Larochelle, and Aaron Courville.
\newblock {HoME}: A household multimodal environment.
\newblock \emph{arXiv preprint arXiv:1711.11017}, 2017.

\bibitem[Chang et~al.(2017)Chang, Dai, Funkhouser, Halber, Nie{\ss}ner, Savva,
  Song, Zeng, and Zhang]{chang2017matterport3d}
Angel Chang, Angela Dai, Thomas Funkhouser, Maciej Halber, Matthias
  Nie{\ss}ner, Manolis Savva, Shuran Song, Andy Zeng, and Yinda Zhang.
\newblock Matterport3d: Learning from {RGB}-d data in indoor environments.
\newblock \emph{International Conference on 3D Vision (3DV)}, 2017.

\bibitem[Chen et~al.(2018)Chen, Shur, Misra, Snavely, and
  Artzi]{chen2018touchdown}
Howard Chen, Alane Shur, Dipendra Misra, Noah Snavely, and Yoav Artzi.
\newblock Touchdown: Natural language navigation and spatial reasoning in
  visual street environments.
\newblock \emph{arXiv preprint arXiv:1811.12354}, 2018.

\bibitem[Cirik et~al.(2018)Cirik, Zhang, and Baldridge]{cirikfollowing}
Volkan Cirik, Yuan Zhang, and Jason Baldridge.
\newblock Following formulaic map instructions in a street simulation
  environment.
\newblock \emph{Visually Grounded Interaction and Language (ViGIL) Workshop,
  NeurIPS}, 2018.

\bibitem[Cueva and Wei(2018)]{cueva2018emergence}
Christopher~J Cueva and Xue-Xin Wei.
\newblock Emergence of grid-like representations by training recurrent neural
  networks to perform spatial localization.
\newblock \emph{International Conference on Learning Representations}, 2018.

\bibitem[Dai et~al.(2017)Dai, Chang, Savva, Halber, Funkhouser, and
  Nie{\ss}ner]{dai2017scannet}
Angela Dai, Angel~X Chang, Manolis Savva, Maciej Halber, Thomas~A Funkhouser,
  and Matthias Nie{\ss}ner.
\newblock {ScanNet}: Richly-annotated 3{D} reconstructions of indoor scenes.
\newblock In \emph{Proceedings of the IEEE Conference on Computer Vision and
  Pattern Recognition (CVPR)}, volume~2, page~10, 2017.

\bibitem[de~Vries et~al.(2018)de~Vries, Shuster, Batra, Parikh, Weston, and
  Kiela]{de2018talk}
Harm de~Vries, Kurt Shuster, Dhruv Batra, Devi Parikh, Jason Weston, and Douwe
  Kiela.
\newblock Talk the walk: Navigating {N}ew {Y}ork {C}ity through grounded
  dialogue.
\newblock \emph{arXiv preprint arXiv:1807.03367}, 2018.

\bibitem[Dosovitskiy et~al.(2017)Dosovitskiy, Ros, Codevilla, L{\'o}pez, and
  Koltun]{dosovitskiy2017carla}
Alexey Dosovitskiy, German Ros, Felipe Codevilla, Antonio L{\'o}pez, and
  Vladlen Koltun.
\newblock Carla: An open urban driving simulator.
\newblock \emph{arXiv preprint arXiv:1711.03938}, 2017.

\bibitem[Espeholt et~al.(2018)Espeholt, Soyer, Munos, Simonyan, Mnih, Ward,
  Doron, Firoiu, Harley, Dunning, Legg, and Kavukcuoglu]{espeholt2018impala}
Lasse Espeholt, Hubert Soyer, Remi Munos, Karen Simonyan, Volodymir Mnih, Tom
  Ward, Yotam Doron, Vlad Firoiu, Tim Harley, Iain Dunning, Shane Legg, and
  Koray Kavukcuoglu.
\newblock Impala: Scalable distributed deep-rl with importance weighted
  actor-learner architectures.
\newblock In \emph{International Conference on Machine Learning (ICML)}, 2018.

\bibitem[Geiger et~al.(2013)Geiger, Lenz, Stiller, and
  Urtasun]{geiger2013vision}
Andreas Geiger, Philip Lenz, Christoph Stiller, and Raquel Urtasun.
\newblock Vision meets robotics: The kitti dataset.
\newblock \emph{The International Journal of Robotics Research}, 32\penalty0
  (11):\penalty0 1231--1237, 2013.

\bibitem[Google(2008)]{protobuf2008}
Google.
\newblock \emph{Protocol Buffers}, 2008.
\newblock URL \url{https://developers.google.com/protocol-buffers/}.
\newblock (accessed 1 March 2019).

\bibitem[Graves et~al.(2017)Graves, Bellemare, Menick, Munos, and
  Kavukcuoglu]{graves2017automated}
Alex Graves, Marc~G Bellemare, Jacob Menick, Remi Munos, and Koray Kavukcuoglu.
\newblock Automated curriculum learning for neural networks.
\newblock In \emph{International Conference on Machine Learning (ICML)}, 2017.

\bibitem[Hochreiter and Schmidhuber(1997)]{hochreiter1997long}
Sepp Hochreiter and J{\"u}rgen Schmidhuber.
\newblock Long short-term memory.
\newblock \emph{Neural computation}, 9\penalty0 (8):\penalty0 1735--1780, 1997.

\bibitem[Kempka et~al.(2016)Kempka, Wydmuch, Runc, Toczek, and
  Ja{\'s}kowski]{kempka2016vizdoom}
Micha{\l} Kempka, Marek Wydmuch, Grzegorz Runc, Jakub Toczek, and Wojciech
  Ja{\'s}kowski.
\newblock Vizdoom: A doom-based ai research platform for visual reinforcement
  learning.
\newblock In \emph{Computational Intelligence and Games (CIG), 2016 IEEE
  Conference on}, pages 1--8. IEEE, 2016.

\bibitem[Kendall et~al.(2015)Kendall, Grimes, and Cipolla]{kendall2015posenet}
Alex Kendall, Matthew Grimes, and Roberto Cipolla.
\newblock Posenet: A convolutional network for real-time 6-dof camera
  relocalization.
\newblock In \emph{Computer Vision (ICCV), 2015 IEEE International Conference
  on}, pages 2938--2946. IEEE, 2015.

\bibitem[Kolve et~al.(2017)Kolve, Mottaghi, Gordon, Zhu, Gupta, and
  Farhadi]{kolve2017ai2}
Eric Kolve, Roozbeh Mottaghi, Daniel Gordon, Yuke Zhu, Abhinav Gupta, and Ali
  Farhadi.
\newblock Ai2-thor: An interactive 3d environment for visual ai.
\newblock \emph{arXiv preprint arXiv:1712.05474}, 2017.

\bibitem[Lample and Chaplot(2017)]{lample_aaai17}
Guillaume Lample and Devendra~Singh Chaplot.
\newblock Playing {FPS} games with deep reinforcement learning.
\newblock In \emph{Proceedings of the Thirty-First {AAAI} Conference on
  Artificial Intelligence}, 2017.

\bibitem[Mirowski et~al.(2017)Mirowski, Pascanu, Viola, Soyer, Ballard, Banino,
  Denil, Goroshin, Sifre, Kavukcuoglu, et~al.]{mirowski2016learning}
Piotr Mirowski, Razvan Pascanu, Fabio Viola, Hubert Soyer, Andrew~J Ballard,
  Andrea Banino, Misha Denil, Ross Goroshin, Laurent Sifre, Koray Kavukcuoglu,
  et~al.
\newblock Learning to navigate in complex environments.
\newblock In \emph{International Conference on Learning Representations
  (ICLR)}, 2017.

\bibitem[Mirowski et~al.(2018)Mirowski, Grimes, Malinowski, Hermann, Anderson,
  Teplyashin, Simonyan, Kavukcuoglu, Zisserman, and
  Hadsell]{mirowski2018learningcityscale}
Piotr Mirowski, Matthew~Koichi Grimes, Mateusz Malinowski, Karl~Moritz Hermann,
  Keith Anderson, Denis Teplyashin, Karen Simonyan, Koray Kavukcuoglu, Andrew
  Zisserman, and Raia Hadsell.
\newblock Learning to navigate in cities without a map.
\newblock \emph{Advances in Neural Information Processing Systems (NeurIPS)},
  2018.

\bibitem[Mnih et~al.(2016)Mnih, Badia, Mirza, Graves, Lillicrap, Harley,
  Silver, and Kavukcuoglu]{mnih2016asynchronous}
Volodymyr Mnih, Adria~Puigdomenech Badia, Mehdi Mirza, Alex Graves, Timothy
  Lillicrap, Tim Harley, David Silver, and Koray Kavukcuoglu.
\newblock Asynchronous methods for deep reinforcement learning.
\newblock In \emph{International Conference on Machine Learning}, pages
  1928--1937, 2016.

\bibitem[Mo et~al.(2018)Mo, Li, Lin, and Lee]{mo2018adobeindoornav}
Kaichun Mo, Haoxiang Li, Zhe Lin, and Joon-Young Lee.
\newblock The {A}dobe{I}ndoor{N}av dataset: Towards deep reinforcement learning
  based real-world indoor robot visual navigation.
\newblock \emph{arXiv preprint arXiv:1802.08824}, 2018.

\bibitem[Moore(1959)]{moore1959shortest}
Edward~F Moore.
\newblock The shortest path through a maze.
\newblock In \emph{Proc. Int. Symp. Switching Theory, 1959}, pages 285--292,
  1959.

\bibitem[Savva et~al.(2017)Savva, Chang, Dosovitskiy, Funkhouser, and
  Koltun]{savva2017minos}
Manolis Savva, Angel~X Chang, Alexey Dosovitskiy, Thomas Funkhouser, and
  Vladlen Koltun.
\newblock Minos: Multimodal indoor simulator for navigation in complex
  environments.
\newblock \emph{arXiv preprint arXiv:1712.03931}, 2017.

\bibitem[Shah et~al.(2018)Shah, Dey, Lovett, and Kapoor]{shah2018airsim}
Shital Shah, Debadeepta Dey, Chris Lovett, and Ashish Kapoor.
\newblock Airsim: High-fidelity visual and physical simulation for autonomous
  vehicles.
\newblock In \emph{Field and Service Robotics}, pages 621--635. Springer, 2018.

\bibitem[Wikipedia(2005)]{equirectangular2005}
Wikipedia.
\newblock \emph{Equirectangular projection}, 2005.
\newblock URL \url{https://en.wikipedia.org/wiki/Equirectangular_projection}.
\newblock (accessed 1 March 2019).

\bibitem[Wu et~al.(2018)Wu, Wu, Gkioxari, and Tian]{wu2018building}
Yi~Wu, Yuxin Wu, Georgia Gkioxari, and Yuandong Tian.
\newblock Building generalizable agents with a realistic and rich 3d
  environment.
\newblock In \emph{European Conference on Computer Vision (ECCV)}, 2018.

\bibitem[Xia et~al.(2018)Xia, Zamir, He, Sax, Malik, and
  Savarese]{xia2018gibson}
Fei Xia, Amir~R Zamir, Zhiyang He, Alexander Sax, Jitendra Malik, and Silvio
  Savarese.
\newblock Gibson {E}nv: Real-world perception for embodied agents.
\newblock In \emph{Proceedings of the IEEE Conference on Computer Vision and
  Pattern Recognition}, pages 9068--9079, 2018.

\bibitem[Yan et~al.(2018)Yan, Misra, Bennnett, Walsman, Bisk, and
  Artzi]{yan2018chalet}
Claudia Yan, Dipendra Misra, Andrew Bennnett, Aaron Walsman, Yonatan Bisk, and
  Yoav Artzi.
\newblock {CHALET}: Cornell house agent learning environment.
\newblock \emph{arXiv preprint arXiv:1801.07357}, 2018.

\bibitem[Zaremba and Sutskever(2014)]{zaremba2014learning}
Wojciech Zaremba and Ilya Sutskever.
\newblock Learning to execute.
\newblock \emph{arXiv preprint arXiv:1410.4615}, 2014.

\bibitem[Zhu et~al.(2017)Zhu, Mottaghi, Kolve, Lim, Gupta, Fei{-}Fei, and
  Farhadi]{zhu_icra2017}
Yuke Zhu, Roozbeh Mottaghi, Eric Kolve, Joseph~J. Lim, Abhinav Gupta,
  Li~Fei{-}Fei, and Ali Farhadi.
\newblock Target-driven visual navigation in indoor scenes using deep
  reinforcement learning.
\newblock In \emph{2017 {IEEE} International Conference on Robotics and
  Automation, {ICRA}}, pages 3357--3364, 2017.

\bibitem[Zuse(1972)]{zuse1972plankalkul}
Konrad Zuse.
\newblock \emph{Der Plankalk{\"u}l}.
\newblock Number~63. Gesellschaft f{\"u}r Mathematik und Datenverarbeitung,
  1972.

\end{thebibliography}

\end{document}